%% file: main.tex
\pdfoutput=1

\documentclass[11pt]{article}

\usepackage[]{EACL2023}

\usepackage{times}
\usepackage{latexsym}

\usepackage[T1]{fontenc}

\usepackage[utf8]{inputenc}

\usepackage{microtype}

\usepackage{subfiles}
\usepackage{graphicx}
\usepackage{caption}
\usepackage{subcaption}
\usepackage{booktabs}
\usepackage{multirow}
\usepackage{xspace}
\usepackage{amsmath}
\usepackage{bm}
\usepackage{enumitem}

\usepackage{xcolor}
\usepackage{soul}
\usepackage{hyperref}

\newif\ifhidecomments
\hidecommentsfalse

\ifhidecomments
    \newcommand{\sam}[1]{}
    \newcommand{\chenhao}[1]{}
    \newcommand{\yiming}[1]{}
    \newcommand{\rosa}[1]{}
    \newcommand{\edit}[1]{}
\else
    \newcommand{\chenhao}[1]{\textcolor{blue}{[#1 ---\textsc{CT}]}}
    \newcommand{\sam}[1]{\textcolor{red}{[#1 ---\textsc{SC}]}}
    \newcommand{\yiming}[1]{\textcolor{purple}{[#1 ---\textsc{YZ}]}}
    \newcommand{\rosa}[1]{\textcolor{orange}{[#1 ---\textsc{Rosa}]}}
    \newcommand{\edit}[1]{\textcolor{violet}{[#1 ---\textsc{Edit}]}}
\fi

\newcommand{\para}[1]{\noindent{\bf #1}}
\newcommand{\figref}[1]{Fig.~\ref{#1}}

\newcommand{\secref}[1]{\S\ref{#1}}
\newcommand{\tableref}[1]{Table~\ref{#1}}

\newcommand{\bert}{BERT\xspace}
\newcommand{\roberta}{RoBERTa\xspace}

\newcommand{\dataaug}{\textsc{Adv.}\xspace}
\newcommand{\dataaugtenx}{\textsc{Adv.-10x}\xspace}
\newcommand{\nodataaug}{\textsc{No Adv.}\xspace}
\newcommand{\dataaugwithnatokens}{\textsc{Adv. + Atk. Sup.}\xspace}
\newcommand{\dataaugwithhuman}{\textsc{Adv. + Human Sup.}\xspace}
\newcommand{\humantokensintable}{\textsc{Human Sup.}\xspace}

\newcommand{\natokens}{\textsc{Non-A}\xspace}

\newcommand{\bc}{BERT\xspace}
\newcommand{\bcaug}{BERT (\dataaug)\xspace}
\newcommand{\br}{BERT rationale\xspace}
\newcommand{\brlr}{human-supervised BERT\xspace}

\newcommand{\brlna}{attack-supervised BERT\xspace}

\newcommand{\qa}{\textsc{query+answer}\xspace}
\newcommand{\hr}{human rationale\xspace}
\newcommand{\vect}[1]{\ensuremath{\bm{#1}}}

\newcommand{\addsent}{\textsc{AddSent}\xspace}

\newcommand{\addsenthuman}{\textsc{AddSentH}\xspace}
\newcommand{\addonesenthuman}{\textsc{AddOneSentH}\xspace}
\newcommand{\addsentsynthetic}{\textsc{AddSentS}\xspace}

\newcommand{\extractor}{rationale extractor\xspace}
\newcommand{\CEL}{\ensuremath{\mathcal{L}_{CE}}}
\newcommand{\multirc}{\textsc{MultiRC}\xspace}
\newcommand{\fever}{\textsc{FEVER}\xspace}
\newcommand{\squad}{\textsc{SQuAD}\xspace}

\newcommand{\nodata}{\text{- - -}}
\newcommand{\attackedS}{Attacked\textsubscript{S}\xspace}
\newcommand{\attackedH}{Attacked\textsubscript{H}\xspace}

\urldef{\repourl}\url{https://github.com/ChicagoHAI/rationalization-robustness}

\title{Learning to Ignore Adversarial Attacks}

\author{Yiming Zhang$^\dagger$ \quad Yangqiaoyu Zhou$^\dagger$ \quad Samuel Carton$^\ddag$ \quad Chenhao Tan$^\dagger$ \\
$^\dagger$ University of Chicago \quad $^\ddag$ University of New Hampshire \\
\small
\texttt{\{yimingz0,zhouy1,chenhao\}@uchicago.edu} \quad \texttt{samuel.carton@unh.edu}
}

\begin{document}
\maketitle
\begin{abstract}
Despite the strong performance of current NLP models, they can be brittle against adversarial attacks. To enable effective learning against adversarial inputs, we introduce the use of rationale models that can explicitly learn to ignore attack tokens. We find that the rationale models can successfully ignore over 90\% of attack tokens. This approach leads to consistent and sizable improvements ($\sim$10\%) 
over baseline models in robustness on three datasets for both \bert and \roberta, 
and also reliably outperforms data augmentation with adversarial examples alone. In many cases, we find that our method is able to 
close the gap between model performance on a clean test set and an attacked test set 
and hence reduce the effect of adversarial attacks.

\end{abstract}

\input{introduction}

\input{related_work}

\input{sections/attacks}

\input{data}



\input{sections/model}
\input{results}

\input{error_analysis}


\input{discussion}

\section*{Limitations}
Our work focuses on improving model robustness by explicitly ignoring
adversarial attacks.
In this work, we only explore a known type of adversarial attack (\addsent),
and the performance of our method against unknown attacks is yet to be
validated.
Since our method uses rationalization as the underlying mechanism for ignoring
tokens, it would take non-trivial work to make our method compatible with
attacks in the form of token removal and flipping.
Finally, we limit our experiments to the domain of QA, where the \addsent attack
is naturally applicable.

\section*{Ethics Statement}
Our work contributes to the line of research that focuses on improving the adversarial
robustness of language models.
We also explore novel ways to integrate human explanations into the training paradigm.
We believe robustness to adversarial attacks is essential to the deployment of
trustworthy models in the wild, and we hope this work brings current research a
step closer to this objective.
To avoid ethical concerns related to over-claiming results, we emphasize in both our concluding discussion and the limitations section that our work builds on the assumption that we know the type of attack and only experiments with \addsent.
Furthermore, our approach tends to increase the computational cost compared to adversarial training both during training and inference.
One should consider the tradeoff between robustness and computation.

\section*{Acknowledgements}
We would like to thank members of the Chicago Human+AI Lab for feedback
on early versions of this work,
and Howard Chen for fruitful discussions on additive adversarial attacks and sharing findings of his concurrent work.
We further thank all anonymous reviewers for their insightful suggestions and comments. 
This work was supported in part by an NSF grant, IIS-2126602, and gifts from Google and Amazon.





\bibliography{custom, sam_robustness_references, yiming_related_work}
\bibliographystyle{acl_natbib}

\appendix
\input{appendix}

\end{document}

%% file: introduction.tex
\section{Introduction}

Adversarial robustness is an important issue in NLP, asking how to proof models against confounding tokens designed to maliciously manipulate model outputs. As such models become more powerful and ubiquitous, research continues to discover surprising vulnerabilities \citep{wallace_universal_2019}, demanding improved robustness methods. 

A common defense method to combat adversarial attacks is adversarial training. 
Given knowledge of attack strategies, 
it constructs synthetic adversarial examples 
to augment clean examples during training \citep{zhang_adversarial_2020}.
Intuitively, the model will \textit{implicitly} learn to ignore attacking tokens and become robust to that type of attack. 
In practice, however, this goal can be challenging through data augmentation alone.

\begin{figure}[t]
    \centering
    \includegraphics[width=0.9\columnwidth]{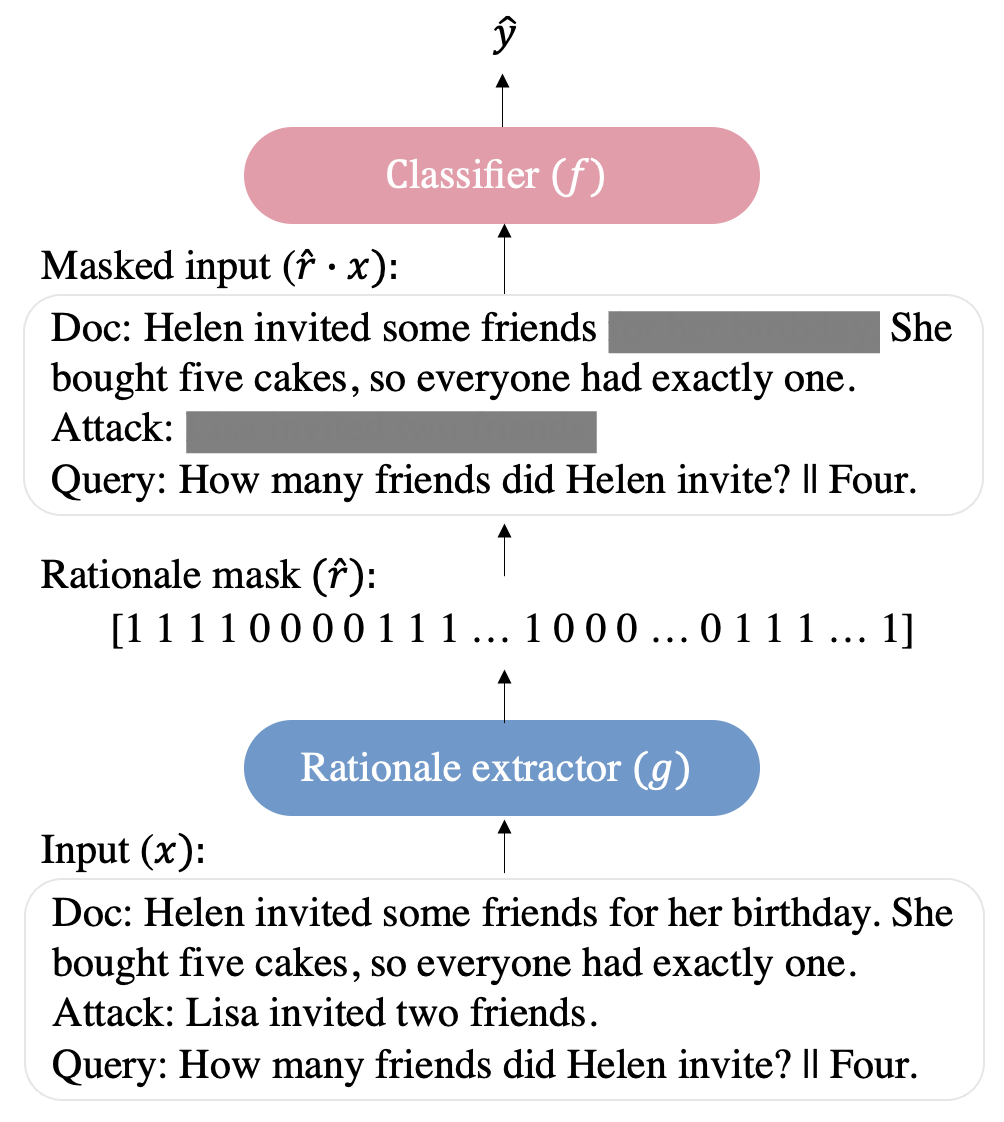}
    \caption{%
    Example illustration of an ideal rationale model that is robust to added attack tokens.
    The rationale extractor filters out the confounding sentence ({\em ``Lisa invited two friends''})
    added by the adversary, and extracts the supporting spans ({\em ``Helen invited some friends''}
    and {\em ``She bought five cakes, so everyone had exactly one''}) to help the model deduce the
    correct answer ({\em ``Four''}).
    }
    \label{fig:model_diagram}
\end{figure}

In this study, we propose a simple yet effective adversarial training schema for additive attacks:
\textit{explicitly} training the model to ignore adversarial tokens. We do this by augmenting the underlying model with 
a rationale extractor \cite{lei-etal-2016-rationalizing} to serve as an input filter, and then training this extractor to ignore attacking tokens as an additional joint objective to overall label accuracy (\figref{fig:model_diagram}). 

In addition to training the extractor to distinguish the attacking/non-attacking token dichotomy, we also explore the utility of human-provided explanations in this regard. In doing so, we ask: does learning from human rationales help the model avoid attending to attacking tokens? 

Fine-tuning BERT \citep{devlin_bert:_2018} and RoBERTa \citep{liu_roberta_2019} 
on 
multiple datasets,
we demonstrate that the additive attack proposed by \citet{jia_adversarial_2017} does reduce model accuracy, and that data augmentation with adversarial examples provides limited benefit in defending these models from this attack in most cases.

Our main results are 
that rationale-style models learn to ignore these attacks more effectively than
only with data augmentation, leading to an improvement of $\sim$10\% 
in accuracy on attacked examples compared to baseline models and an advantage of 
$2.4$\%
over data augmentation alone, mostly recovering clean test performance.
While human explanations may potentially improve the interpretability of these models, they are of limited use in improving this defense even further. 

In summary, we offer three main contributions:


\setlist{nolistsep}
\begin{itemize}[leftmargin=*]
    \item We show that explicitly training an extractive rationale layer to ignore attack tokens is more effective 
    than implicitly training a model via data augmentation with adversarial examples.
    \item We assess whether human-annotated rationales augment this defense, showing that they have only a limited benefit. 
    \item We conduct an in-depth error analysis of differences between models, explaining some of the patterns we observe in our main results. 
\end{itemize}

Our code is available at \repourl.

%% file: related_work.tex
\section{Related Work}


We build on prior work on adversarial robustness and learning from explanations.

\para{Adversarial robustness.}
\input{sections/adversarial-robustness}

\para{Learning from explanations.}
Recent work has sought to collect datasets of human-annotated explanations, often in the form of binary \textit{rationales}, in addition to class labels \citep{deyoung_eraser_2019,wiegreffe_teach_2021}, and to use these explanations as additional training signals to improve model performance and robustness, sometimes also known as \textit{feature-level feedback} \citep{hase_when_2021,beckh_explainable_2021}.

An early work is \citet{zaidan_using_2007}, which uses human rationales as constraints on an SVM. More recently, \citet{ross_right_2017} uses human rationales to penalize neural net input gradients showing benefits for out-of-domain generalization, while \citet{erion_improving_2021} use a similar method based on ``expected gradients'' to produce improvements in in-domain test performance in certain cases.  \citet{katakkar_practical_2021} evaluate feature feedback for two attention-style models, finding, again, gains in out-of-domain performance, while \citet{han_influence_2021} use influence functions \citep{koh_understanding_2017} to achieve a similar outcome. Where our study differs from most previous work is in using feature feedback for adversarial rather than out-of-domain robustness. A concurrent work by \citet{chen2022RationaleRobustness} uses rationalization
to improve robustness. The proposed method is similar to our work, but we explore supervision with attack tokens and achieve stronger robustness to additive attacks.

%% file: sections/adversarial-robustness.tex
Adversarial attacks against NLP models seek to maliciously manipulate model output by perturbing model input. \citet{zhang_adversarial_2020} present a survey of both attacks and defenses.
Example attacks include character-level manipulations \citep{gao_black-box_2018,li_textbugger_2019}, input removal \citep{li_understanding_2017, feng_pathologies_2018}, synonym substitutions \citep{ren_generating_2019}, and language model-based slot filling \citep{li_bert-attack_2020,garg_bae_2020,li_contextualized_2021}.
A distinction in attack types is whether the attack requires access to the model \citep{ebrahimi_hotflip_2018,yoo_towards_2021,wallace_universal_2019} or not \citep{alzantot_generating_2018,jin_is_2020}.
TextAttack \citep{morris_textattack_2020} is a framework and collection of attack implementations.
Our work focuses on the \addsent attack proposed by \citet{jia_adversarial_2017} in reading comprehension.
As interest in adversarial attacks has increased, so has interest in developing models robust to these attacks. A popular defense method is adversarial training via data augmentation, first proposed by \citet{szegedy_intriguing_2014} and employed by \citet{jia_adversarial_2017} to bring their model \textit{almost} back to clean test performance. A recent example in this vein is \citet{zhou_defense_2020}, which proposes Dirichlet Neighborhood Ensemble as a means for generating dynamic adversarial examples during training. Another popular approach is knowledge distillation \citep{papernot_distillation_2016}, which trains an intermediate model to smooth between the training data and the final model. 
Our work explores a new direction that explicitly learns to ignore  attacks.




%% file: sections/attacks.tex
\section{Adversarial Attacks and Datasets}


In this paper, we focus on model robustness against the \addsent additive attack proposed by \citet{jia_adversarial_2017}. The attack is designed for
reading comprehension: consider each instance as a tuple of document, query, and label $(d, q, y)$, where $y$ indicates whether the query is supported by the document.
The attack manipulates the content of the query to form an attack sentence ($A$) and adds $A$ to the document to confuse the model. 
Specifically, \addsent proceeds as follows:


\setlist{nolistsep}

\begin{enumerate}[leftmargin=*]
    \item[1.] We modify the query $q$ by converting all named entities and numbers to their nearest neighbor in the GloVe embedding space \citep{pennington_glove_2014}. We flip all adjectives and nouns to their antonyms using WordNet \citep{miller_wordnet_1995} and yield a mutated query $\hat{q}$. If we fail to mutate the query due to not being able to find matching named entities or antonyms of adjectives and nouns, we skip the example.
    
    \item[2.] We convert the mutated query $\hat{q}$ into an adversarial attack $A$ using CoreNLP \citep{manning_stanford_2014} constituency parsing, under a set of about 50 rules enumerated by \citet{jia_adversarial_2017}. This step converts it into a factual statement that resembles but is not semantically related to the original query $q$.
    
    \item[3.] The adversarial attack $A$ is inserted at 
    a random location within the original document and leads to a new tuple $(d', q, y)$.\footnote{We experimented with variants of inserting only at the beginning or the end. The results are qualitatively similar, so we only report random in this paper.}
\end{enumerate}

The key idea behind the \textsc{AddSent} attack is that the mutations alter the semantics of the query by mutating the named entities and numbers,
so that the attack contains words or phrases that are likely confusing to the model without changing the true semantics of the input. An example of the \addsent attack is given above.








\begin{figure}[t]
\small
\centering
\fbox{\begin{minipage}{\linewidth}
\textbf{Query} $q$:

FC Bayern Munich was founded in 2000.

\textbf{Mutated Query} $\hat{q}$:

DYNAMO Leverkusen Cologne was founded in 1998.

\textbf{Modified Document} $d'$

\dots has won 9 of the last 13 titles. DYNAMO Leverkusen Cologne was founded in 1998. They have traditional local rivalries with \dots
\end{minipage}}

\caption{An example of the \addsent attack.}
\label{fig:addsent_example}
\end{figure}

The original approach includes an additional step of using crowdsourced workers to filter ungrammatical sentences.
We do not have access to this manual validation process in all datasets.
Occasionally, \addsent generates ungrammatical attacks
but it nevertheless proves empirically effective in reducing the performance of our models.



%% file: data.tex

\para{Datasets.}
To evaluate our hypotheses on learning to ignore adversarial attacks, we  
train and evaluate models on the  Multi-Sentence Reading Comprehension \cite[\multirc;][]{khashabi_looking_2018} and Fact Extraction and VERification \cite[\fever;][]{thorne_fever_2018}  datasets. Both are reading comprehension datasets, compatible with the \addsent attack.
For \multirc, the query consists of a question and potential answer about the document, labeled as true or false, while for \fever it is a factual claim about the document labeled as ``supported'' or ``unsupported''. 
Both datasets include \textit{human rationales}, 
indicating which tokens 
are pertinent to assessing the query. Table \ref{tab:data} summarizes their basic statistics.

\begin{table}[t]
\small
\centering
\begin{tabular}{@{}llll@{}}
\toprule
Dataset                 & \begin{tabular}[c]{@{}l@{}}Text \\ length\end{tabular} & \begin{tabular}[c]{@{}l@{}}Rationale \\ length\end{tabular} & \begin{tabular}[c]{@{}l@{}} Total \\ size \end{tabular}   \\ \midrule
\multirc & 336.0                                                    & 52.0                                                          & 32,088 \\
\fever   & 335.9                                                    & 47.0                                                          & 110,187 \\
\squad   & 119.8                                                    & ------                                                        & 87,599 \\
\bottomrule
\end{tabular}
    \caption{Basic statistics of \multirc, \fever, and \squad.}
    \label{tab:data}
\end{table}

In modeling these two datasets, we follow standard practice in appending the query
to the end of the document with [SEP] tokens. 
We use train/validation/test splits prepared by the ERASER dataset collection \citep{deyoung_eraser_2019}. 
Because we are interested in relative differences between training regimes rather than absolute performance, we subsample the \fever training set to 25\% so that it is comparable to \multirc for the sake of training efficiency.

Directly applying the synthetic \addsent attack to \multirc and \fever leads to occasionally ungrammatical adversarial examples due to incorrectly applied conversion heuristic or errors in constituency parsing. 
To alleviate this concern, we further evaluate on \squad \citep{rajpurkar_squad_2016}, for which \citeauthor{jia_adversarial_2017} provide an \addsent-attacked {\em evaluation} set that is re-written and approved by human workers.
However, this dataset does not have human rationales. 
We again use the train/validation/test splits provided by \citeauthor{jia_adversarial_2017} in our experiments.



%% file: sections/model.tex
\input{tables/supervision_table.tex}

\begin{table}[t]
    \centering
    \small
    \begin{tabular}{ll}
    \toprule
    Data augmentation? & Rationale? \\
    \midrule
    \multirow{2}{*}{\shortstack{No data \\ augmentation}} & None \\     
    & Human (\humantokensintable) \\
    \midrule
    \multirow{3}{*}{\shortstack{Augmented with \\ attack data (Adv.)}} & None \\     
    & Non-attack (\dataaugwithnatokens)\\
    & Human (\dataaugwithhuman) \\
    \bottomrule
    \end{tabular}
        \caption{Summaries of rationale model setups.}
        \label{tab:models}
\end{table}

\section{Modeling}


Our study assesses whether adding an explicit 
\extractor
to a 
model and training it to ignore attack tokens results in a more effective defense than simply adding attacked examples to the training set. This comparison results in several combinations of model architectures and training regimes.

We denote each training instance as $(\vect{x}, \vect{r}, y)$: a text sequence $\vect{x}$ consisting of the concatenated document and query, a ground-truth binary rationale sequence $\vect{r}$, and a binary label $y$.

\para{Baseline models and training.}
We use BERT \citep{devlin_bert:_2018} and RoBERTa \citep{liu_roberta_2019} as basis models. 
In the baseline training condition we fine-tune these models as normal, evaluating them on both the original test set and a version of the test set where each item has been corrupted with the \addsent attack described above. 
We denote this condition as ``\nodataaug''

In the baseline adversarial training via data augmentation condition (denoted \dataaug), we add \addsent-attacked versions of each training example to the training set on a one-to-one basis, allowing the model to train for the presence of such attacks. This represents a fairly standard baseline defense in the literature \citep{zhang_adversarial_2020}.

Following prior adversarial robustness literature \citep{jia_certified_2019}, we also consider a stronger baseline by augmenting the training set with $K$ perturbed examples for each training example. For our main experiments, we use $K = 10$. This setting (denoted \dataaugtenx) measures whether the baseline method implicitly adapts to the \addsent attack when abundant signal is provided.

\para{Rationale model.}
To lend the baseline model an extractor capable of filtering out confounding tokens, we use the rationale model proposed by \citet{lei-etal-2016-rationalizing}. It comprises a rationale extractor $g$ and a label predictor $f$ (Fig. \ref{fig:model_diagram}). The rationale extractor generates a binary predicted rationale $\hat{\vect{r}}$, which is applied as a mask over the input to the predictor via masking function $m$, producing a predicted label:
\begin{equation}
\begin{aligned}
& g(\mathbf{x}) \rightarrow \hat{\vect{r}} \\
& f(m(\mathbf{x}, \hat{\vect{r}})) \rightarrow \hat{y}
\end{aligned}
\end{equation}
The two components are trained together to optimize predicted label accuracy as well as loss associated with the predicted rationale. In an unsupervised scenario, this loss punishes the norm of the predicted rationale, encouraging sparsity on the (heuristic) assumption that a sparse rationale is more interpretable. In this study, we rather consider the supervised scenario, where we punish $\hat{\vect{r}}$'s error with respect to a ground-truth rationale $\vect{r}$. However, we find empirically that the rationale sparsity objective is useful in combination with the rationale supervision objective, leading to the following joint objective function using cross-entropy loss $\CEL$ with hyperparameter weights $\lambda_1$ and $\lambda_2$:
\begin{equation}
 \CEL(\hat{y},y) + 
 \lambda_1 \CEL(\hat{\vect{r}},\vect{r}) + 
 \lambda_2 ||\hat{\vect{r}}||.
\end{equation}




\paragraph{Adversarial training with rationale supervision.}
To introduce rationale supervision, we augment the training set with attacked examples on a one-to-one basis with original examples, similar to adversarial training.
Moreover, we can change the ground-truth rationale to reflect the desired behavior for the model. We consider two options for this new ground-truth $\vect{r}$: (1) a binary indicator of whether a token is adversarial or not (\dataaugwithnatokens), and (2) the 
human-annotated rationale (\dataaugwithhuman), which also filters adversarial tokens. 
Table~\ref{tab:supervision} contains an example illustrating the distinction between
\dataaugwithnatokens and \dataaugwithhuman

Table~\ref{tab:models} summarizes all the combinations of setups that we use in our study.
For each of these setups, we test one rationale model using independent BERT modules for $g$ and $f$, and one using independent \roberta modules for both. 
We present additional implementation details in Appendix~\ref{appendix:implementation}.

Taken together, these conditions address our three research questions:
(1) Is adversarial training via rationale supervision more effective than via attacked examples?
(2) Does training the model to emulate human explanation make it intrinsically more robust to attacks?
(3) Do human explanations improve upon adversarial training with non-attack tokens as rationale supervision?

%% file: tables/supervision_table.tex
\definecolor{human}{RGB}{244,236,164}
\definecolor{attack}{RGB}{238,186,180}
\definecolor{supervision}{RGB}{163,189,219}
\definecolor{ignore}{RGB}{180,180,180}
\DeclareRobustCommand{\hlhuman}[1]{{\sethlcolor{human}\hl{#1}}}
\DeclareRobustCommand{\hlattack}[1]{{\sethlcolor{attack}\hl{#1}}}
\DeclareRobustCommand{\hlsupervision}[1]{{\sethlcolor{supervision}\hl{#1}}}
\DeclareRobustCommand{\hlignore}[1]{{\sethlcolor{ignore}\hl{#1}}}
\DeclareRobustCommand{\hlnull}[1]{{\sethlcolor{white}\hl{#1}}}

\begin{table*}[ht!]
    \centering
\footnotesize
\begin{tabular}{@{}p{0.31\linewidth} p{0.31\linewidth} p{0.31\linewidth}@{}}
\toprule
\multicolumn{1}{c}{\textbf{Human rationale \& attack}} & \multicolumn{1}{c}{\textbf{\dataaugwithnatokens}} & \multicolumn{1}{c}{\textbf{\dataaugwithhuman}}    \\ \midrule
\hlnull{... and 18 national cups.}
\hlhuman{FC Bayern was founded in 1900 by 11 football players, led by Franz John.}
\hlnull{Although Bayern won ... European Cup three times in a row (1974 -- 1976).}
\hlattack{DYNAMO Leverkusen Cologne was founded in 1998.}
\hlnull{Overall , Bayern has reached ten European ...}
&
\hlsupervision{... and 18 national cups.}
\hlsupervision{FC Bayern was founded in 1900 by 11 football players, led by Franz John.}
\hlsupervision{Although Bayern won ... European Cup three times in a row (1974 -- 1976).}
\hlignore{DYNAMO Leverkusen Cologne was founded in 1998.}
\hlsupervision{Overall , Bayern has reached ten European ...}
&
\hlignore{... and 18 national cups.}
\hlsupervision{FC Bayern was founded in 1900 by 11 football players, led by Franz John.}
\hlignore{Although Bayern won ... European Cup three times in a row (1974 -- 1976).}
\hlignore{DYNAMO Leverkusen Cologne was founded in 1998.}
\hlignore{Overall , Bayern has reached ten European ...}
\\ \bottomrule
\end{tabular}
\caption{An example from \fever illustrating different modes of adversarial training with rationale supervision.
Human rationales are colored \hlhuman{yellow}, and attack tokens are colored \hlattack{red}.
Rationale models are supervised to extract the \hlsupervision{blue} tokens,
while ignoring the \hlignore{gray} tokens.
}
\label{tab:supervision}
\end{table*}

%% file: results.tex
\section{Experimental Setup and Results}

We start by describing our experimental setup and evaluation metrics.
We then investigate model performance with different training regimes and conduct an in-depth error analysis.

\input{tables/big-table}
\input{tables/f1}
\subsection{Experimental Setup}


Our study compares whether rationale-style models are better at learning to explicitly ignore adversarial tokens than standard models via adversarial training. As we describe above, we train three variants of the standard classification model (\nodataaug,  \dataaug, \dataaugtenx), and three variants of the rationale model (\dataaugwithnatokens, \humantokensintable, \dataaugwithhuman). 

Exploring these 6 architecture/training combinations for three datasets (\multirc, \fever, 
and \squad%
) and two underlying models (\bert and \roberta), we report results from all trained models in Table \ref{tab:accuracy_table}. We report relevant metrics on both the clean test set and the attacked test set 
for each model.
For \multirc and \fever, the metric we use is accuracy. Since \squad is a span extraction task, we report the Span F1 score instead.
Performance on the attacked test set is our key measure of robustness.


Additionally, for the rationale models, we report 
the mean percentage of attack and non-attack tokens included in each predicted rationale, two metrics that help explain our accuracy results. 
The mean percentage of attack tokens included in the predicted rationale indicates the effectiveness of ignoring attack tokens: the lower the better.

\subsection{Main Results} \label{sec:main_results}
We focus our analysis on three questions:

\setlist{nolistsep}
\begin{enumerate}[leftmargin=*]
    \item Does adversarial rationale supervision on augmented data improve robustness over adversarial data augmentation alone?
    \item Does human rationale supervision improve adversarial robustness over a standard model?
    \item Does the addition of human rationales to adversarial training further improve robustness?
\end{enumerate}

Table \ref{tab:accuracy_table} summarizes the main results of the paper, showing the accuracy of each combination of architecture, training regime, underlying model and dataset. Looking at the attacked versus clean test set performance for the standard model, we see that \textbf{the \addsent attack is effective}, 
reducing accuracy on \multirc ($\sim$6\%), \fever ($\sim$10\%), and \squad ($\sim$12-24\%).




\para{Adversarial rationale supervision (\dataaugwithnatokens).} 
Rationale models provide an interface for explicitly supervising the model to ignore attack tokens. 
Our {\em key} question is whether they can be used to improve the effectiveness of adversarial training.
We first discuss the effect of data augmentation and then show that rationale models are indeed more effective at ignoring attack tokens.

\textit{Data augmentation with adversarial examples works, mostly.} In almost all cases, data augmentation does result in improved performance on the attacked test set, 
improving +5.9\% (\fever) and +17.6\% (\squad) for \bert, as well as +6.4\% (\multirc), +9.7\% (\fever), and +9.4\% (\squad) for \roberta. 
The exception is \bert on \multirc, where it causes a decrease of -1.0\%. However, in only one case out of six does data augmentation with adversarial examples bring the model back to clean test performance (\roberta on \multirc, +0.3\%).

Surprisingly, \bert on \multirc is the only scenario where the \dataaugtenx augmentation significantly improves attack accuracy (4.3\% improvement over \dataaug). In all the other cases, 
adding more adversarial examples does not improve robustness and even leads to a 3.5\% drop in \squad for \roberta.
This result demonstrates that \bert and \roberta may not learn from adversarial examples alone.

\textit{Adversarial rationale supervision improves on adversarial training baselines in all cases.} 
We see an improvement of +4.6\% for \bert on \multirc, +2.9\% for \bert on \fever, +2.7\% for \bert on \squad, +2.2\% for \roberta on \multirc, +0.7\% for \roberta on \fever, and +1.0\% for \roberta on \squad (2.4\% on average). 
For the one case where adversarial data augmentation recovered clean test performance (\roberta on \multirc), adversarial rationale supervision actually improves clean test performance by +2.5\%.

The effectiveness of \dataaugwithnatokens is even more salient if we compare with \nodataaug on attacked test: 
3.6\%, 8.8\%, and 20.3\% for \bert on \multirc, \fever, and \squad, 8.6\%, 10.4\%, and 10.4\% for \roberta on \multirc, \fever, and \squad (10.4\% on average). 

The above findings remain true even when we compare our methods against the theoretically stronger baseline of \dataaugtenx, where the training dataset is augmented with 10 perturbed examples for every training example. Our models trained with adversarial rationale supervision outperforms \dataaugtenx across all datasets and models, and our best model outperforms the \dataaugtenx baseline by 3.3\%
on average. This result highlights both the efficiency and the effectiveness of our method: with adversarial rationale supervision, \bert and \roberta achieve greater defense against the \addsent attack using 
10\% of the adversarial examples.

Interestingly, the adversarially-supervised rationale model demonstrates a strong ability to generalize knowledge learned from synthetic attacks to tune out human-rewritten attacks (+20.3\% on \squad; recall we do not have human-rewritten attacks during training), indicating the potential of our method in a real-world scenario.

Table \ref{tab:percentage_inclusion_table} explains this success. The adversarially-supervised rationale model includes 6\% or fewer attacking tokens on \multirc and \fever, indicating that it did largely succeed in learning to occlude these tokens for the predictor. Additionally, both \bert and \roberta rationale models are able to tune out most human-generated attack tokens, ignoring over 70\% of attack tokens while keeping 99\% of the original text for both models.

\para{Effect of human rationale supervision alone (\humantokensintable).}
We find 
mixed evidence for whether human rationale supervision alone improves adversarial robustness. For BERT on \multirc and \roberta on \fever, human rationale outperforms the standard classification model, but the opposite occurs for the other two model/dataset combinations. 

Table \ref{tab:percentage_inclusion_table} contextualizes this mixed result: the rationale model supervised solely on human rationales includes 60.0\% to 92.4\% of attack tokens in its rationale (compared to between 8.2\% and 17.8\% of non-attack tokens), indicating that it is largely fooled by the \addsent attack into exposing the predictor to attack tokens. 

This result may be explained by the fact that human rationales for these datasets identify the part of the document that pertains particularly to the query, while the \addsent attack crafts adversarial content with a semantic resemblance to that same query. Hence, it is understandable that human rationale training  would not improve robustness. 

\para{Human and adversarial rationale supervision (\dataaugwithhuman).}
Although human rationales alone may not reliably improve model robustness, a final question is whether human rationales can serve as a useful addition to adversarial training. Does training the model to both ignore adversarial tokens and emulate human explanations further improve robustness against the \addsent attack?  

In two out of four cases, the performance of \dataaugwithhuman is equal to that of \dataaugwithnatokens Only for \bert on \multirc does \dataaugwithhuman result in an improvement, being the only configuration 
that brings performance back to that of clean test for that model, and dataset. For \roberta on \multirc, it actually weakens attacked test performance. 

While these results are mixed, Table \ref{tab:percentage_inclusion_table} shows that the model does at least achieve this result at a much lower included percentage of non-attack  tokens ($\sim$20\% vs. $>$95\%), a concession toward model interpretability. 

Overall, our results suggest that human rationales have limited effect in defending against adversarial attacks, but can be important in developing sparse (and potentially interpretable) models.

%% file: tables/big-table.tex

\begin{table*}[t] 
\centering
\small
\begin{tabular}{@{}lllrrrrrr@{}}
\toprule
\multicolumn{1}{c}{\multirow{2}{*}{Model}} &
  \multicolumn{1}{c}{\multirow{2}{*}{Architecture}} &
  \multicolumn{1}{c}{\multirow{2}{*}{Training}} &
  \multicolumn{2}{c}{\multirc (Acc.)} &
  \multicolumn{2}{c}{\fever (Acc.)} &
  \multicolumn{2}{c}{\squad (Span F1)} \\ \cmidrule(l){4-9} 
\multicolumn{1}{c}{} &
  \multicolumn{1}{c}{} &
  \multicolumn{1}{c}{} &
  Clean &
  \attackedS &
  Clean &
  \attackedS &
  Clean &
  \attackedH \\ \midrule
\multirow{5}{*}{\bert} &
  \multirow{3}{*}{Standard} &
  \nodataaug &
  68.6 &
  62.6 &
  88.2 &
  78.9 &
  86.4 &
  62.8 \\
 &
   &
  \dataaug &
  67.3 &
  61.6 &
  \textbf{88.5} &
  84.8 &
  86.0 &
  80.4 \\
 &
   &
  \dataaugtenx &
  66.2 &
  65.9 &
  86.3 &
  84.5 &
  82.2 &
  78.0 \\ \cmidrule(l){2-9} 
 &
  \multirow{3}{*}{Rationale} &
  \dataaugwithnatokens &
  69.6 &
  66.2 &
  87.1 &
  \textbf{87.7} &
  \textbf{86.5} &
  \textbf{83.1} \\
 &
   &
  \humantokensintable &
  70.0 &
  64.4 &
  88.0 &
  76.7 &
  \nodata &
  \nodata \\
 &
   &
  \dataaugwithhuman &
  \textbf{70.5} &
  \textbf{69.4} &
  87.5 &
  87.5 &
  \nodata &
  \nodata \\
  \midrule
\multirow{5}{*}{\roberta} &
  \multirow{3}{*}{Standard} &
  \nodataaug &
  82.6 &
  76.5 &
  93.5 &
  83.0 &
  93.2 &
  81.0 \\
 &
   &
  \dataaug &
  84.4 &
  82.9 &
  93.2 &
  92.7 &
  92.9 &
  90.4 \\
 &
   &
  \dataaugtenx &
  83.5 &
  82.1 &
  93.5 &
  93.2 &
  89.9 &
  86.9 \\ \cmidrule(l){2-9} 
 &
  \multirow{3}{*}{Rationale} &
  \dataaugwithnatokens &
  \textbf{85.2} &
  \textbf{85.1} &
  93.4 &
  \textbf{93.4} &
  \textbf{93.3} &
  \textbf{91.4} \\
 &
   &
  \humantokensintable &
  84.0 &
  74.9 &
  \textbf{94.1} &
  85.7 &
  \nodata &
  \nodata \\
 &
   &
  \dataaugwithhuman &
  85.0 &
  82.5 &
  93.4 &
  \textbf{93.4} &
  \nodata &
  \nodata \\
  \bottomrule
\end{tabular}
\caption{
Model performance on clean and attacked test sets for \multirc, \fever, and \squad.
\attackedS are synthetic attacks produced by \addsent, and \attackedH~are attacks generated by human workers. We vary the level of augmentation for the standard classification models (\nodataaug, \dataaug, \dataaugtenx). For rationale models, we control for the presence of adversarial training data and the type of rationale supervision:
\dataaugwithnatokens treats non-attack tokens as rationale, and \humantokensintable does not use adversarial training. 
Rationale models outperform the baseline classifiers across all attacked datasets.
} 
\label{tab:accuracy_table} 
\end{table*}

%% file: tables/f1.tex
\begin{table*}[]
\centering
\small
\begin{tabular}{@{}llrrrrrr@{}}
\toprule
\multicolumn{1}{c}{\multirow{2}{*}{Model}} &
  \multicolumn{1}{c}{\multirow{2}{*}{Training}} &
  \multicolumn{2}{c}{\multirc} &
  \multicolumn{2}{c}{\fever} &
  \multicolumn{2}{c}{\squad} \\ \cmidrule(l){3-8} 
\multicolumn{1}{c}{} &
  \multicolumn{1}{c}{} &
  \multicolumn{1}{l}{Attack \%} &
  \multicolumn{1}{l}{\natokens \%} &
  \multicolumn{1}{l}{Attack \%} &
  \multicolumn{1}{l}{\natokens \%} &
  Attack \% &
  \natokens \% \\ \midrule
\multirow{3}{*}{\bert}    & \dataaugwithnatokens & 1.4  & 98.4 & 0.2  & 96.7 & 27.8                   & 99.7                   \\
                                         & \humantokensintable  & 87.5 & 8.2  & 66.7 & 17.8 & \nodata & \nodata \\
                                         & \dataaugwithhuman    & 9.5  & 14.4 & 0.5  & 24.4 & \nodata & \nodata \\ \midrule
\multirow{3}{*}{\roberta} & \dataaugwithnatokens & 6.0  & 96.7 & 0.9  & 95.8 & 16.1                   & 99.0                   \\
                                         & \humantokensintable  & 92.4 & 12.6 & 60.0 & 12.2 & \nodata & \nodata \\
                                         & \dataaugwithhuman    & 32.1 & 15.6 & 0.1  & 23.0 & \nodata & \nodata \\ 
                        \bottomrule 
\end{tabular}
\caption{Percentage of attack and non-attack (\natokens) tokens \textit{included} in the predicted rationales. Lower is better for attack tokens.
Arguably, a lower percentage of non-attack tokens is also better as it improves interpretability.}
\label{tab:percentage_inclusion_table}
\end{table*}

%% file: error_analysis.tex
\input{tables/example_table}

\subsection{Error Analysis}

To better understand the behavior of the models, we examine mistakes from BERT compared to explicitly training a \extractor on \multirc. 
We start with a qualitative analysis of example errors, and then discuss general trends, especially on why human rationales only provide limited benefits over \dataaugwithnatokens
More in-depth analyses can be found in the appendix for space reasons, including a Venn diagram of model mistakes.

\para{Qualitative analysis.}
We look at example errors of \dataaug to investigate attacks that are confusing even after adversarial augmentation. 
Table~\ref{tab:example_table} shows example outputs of the rationale models based on either non-attack tokens or human rationales. 

Example 1
shows a case where models with explicit rationale extractors ignore attacks more effectively than \dataaug
In the attack sentence, ``{\em tete didn't stay in}'' is highly similar to the query, so a model likely predicts True if it uses the attack information.
In comparison, both rationale models ignore the attack in label prediction, which enables them to make correct predictions.

Example 2 demonstrates that
\dataaugwithhuman makes mistakes when it fails to include crucial information in rationales while avoiding attack tokens.
\dataaugwithhuman predicts the wrong label because it misses information for the number of friends in its rationale.
\dataaugwithnatokens gets this example correct because it can both ignore the attack and include the necessary information.



Finally, Example 3 shows an example where
\dataaugwithhuman is better than \dataaugwithnatokens when generating rationales to ignore noises.
\dataaugwithhuman includes attacks in rationale, but it is still able to predict the label because the attack is not confusing given the selected rationale. 
The generated rationale helps \dataaugwithhuman to avoid unnecessary information that may confuse the model. 
For example, the sentence with ``picts'' could confuse the model to predict True.
On the other hand, \dataaugwithnatokens gets this example wrong, despite occluding all attack tokens.

More generally, we find that \textit{\dataaugwithhuman tends to have high false negative rates}. 
When \dataaugwithhuman fails to extract good rationales, it tends to predict False 
due to missing information from the rationale.
In contrast, \dataaugwithnatokens rarely occludes necessary information, so it does not suffer from the same issue.

\textit{\dataaugwithnatokens is better than \dataaugwithhuman when human rationales are denser and passage length is longer (see Table \ref{tab:brlr-vs-brlna} in the appendix).} 
We observe that denser human rationales usually comprise evidence from different parts of the passage.
Since \dataaugwithnatokens predicts almost all non-attack tokens as rationale, they have higher \hr recall (98.6\%) than 
\dataaugwithhuman
(57.6\%). Thus, \dataaugwithnatokens generates higher quality rationales when human rationales are dense.
Similarly, long passages prove difficult for \dataaugwithhuman

In summary, these analyses highlight the challenges of learning from human rationales: it requires precise occlusion of irrelevant tokens while keeping valuable tokens, and must account for variance in human rationale and input lengths.
These challenges partly explain the limited benefit of \dataaugwithhuman over \dataaugwithnatokens




%% file: tables/example_table.tex
\begin{table*}[!ht]
\footnotesize
\begin{tabular}{@{}p{0.31\linewidth} p{0.31\linewidth} p{0.31\linewidth}@{}}
\toprule
\multicolumn{1}{c}{\textbf{Human rationale \& attack}} & \multicolumn{1}{c}{\textbf{\dataaugwithnatokens}} & \multicolumn{1}{c}{\textbf{\dataaugwithhuman}}    \\ \midrule
\multicolumn{3}{c}{\textbf{(A) Example 1}, true label: False}     \\ \midrule
\definecolor{highlight}{RGB}{244,236,164 }\sethlcolor{highlight}\hl{[CLS]}\xspace\xspace\xspace...\xspace\xspace 
\hl{in}\hl{ }\hl{may}\hl{ }\hl{1904}\hl{ }\hl{,}\hl{ }\hl{the}\hl{ }\hl{couple}\hl{ }\hl{'}\hl{ }\hl{s}\hl{ }\hl{first}\hl{ }\hl{son}\hl{ }\hl{,}\hl{ }\hl{hans}\hl{ }\hl{albert}\hl{ }\hl{einstein}\hl{ }\hl{,}\hl{ }\hl{was}\hl{ }\hl{born}\hl{ }\hl{in}\hl{ }\hl{bern}\hl{ }\hl{,}\hl{ }\hl{switzerland}\hl{ }\hl{.}
\hl{ }\hl{their}\hl{ }\hl{second}\hl{ }\hl{son}\hl{ }\hl{,}\hl{ }\hl{eduard}\hl{ }\hl{,}\hl{ }\hl{was}\hl{ }\hl{born}\hl{ }\hl{in}\hl{ }\hl{zurich}\hl{ }\hl{in}\hl{ }\hl{july}\hl{ }\hl{1910}\hl{ }\hl{.}\hl{ }\hl{in}\hl{ }\hl{1914}\hl{ }\hl{,}\hl{ }\hl{the}\hl{ }\hl{couple}\hl{ }\hl{separated}\hl{ }\hl{;}\hl{ }\hl{einstein}\hl{ }\hl{moved}\hl{ }\hl{to}\hl{ }\hl{berlin}\hl{ }\hl{and}\hl{ }\hl{his}\hl{ }\hl{wife}\hl{ }\hl{remained}\hl{ }\hl{in}\hl{ }\hl{zurich}\hl{ }\hl{with}\hl{ }\hl{their}\hl{ }\hl{sons}\hl{ }\hl{.} they divorced on 14 february 1919 , having lived apart for five years .\xspace\xspace\xspace...\xspace\xspace \definecolor{highlight}{RGB}{238,186,180} \sethlcolor{highlight} \hl{a}\hl{ }\hl{-}\hl{ }\hl{te}\hl{te}\hl{ }\hl{did}\hl{ }\hl{n}\hl{ }\hl{'}\hl{ }\hl{t}\hl{ }\hl{stay}\hl{ }\hl{in}\hl{ }\hl{basel}\hl{ }\hl{after}\hl{ }\hl{charles}\hl{ }\hl{and}\hl{ }\hl{ho}\hl{ub}\hl{en}\hl{ }\hl{separated}\hl{ }\hl{.} \xspace\xspace\xspace...\xspace\xspace \definecolor{highlight}{RGB}{244,236,164} \sethlcolor{highlight}  \hl{[SEP]}\hl{ }\hl{who}\hl{ }\hl{did}\hl{ }\hl{n}\hl{ }\hl{'}\hl{ }\hl{t}\hl{ }\hl{stay}\hl{ }\hl{in}\hl{ }\hl{zurich}\hl{ }\hl{after}\hl{ }\hl{albert}\hl{ }\hl{and}\hl{ }\hl{mari}\hl{c}\hl{ }\hl{separated}\hl{ }\hl{?}\hl{ }\hl{|}\hl{ }\hl{|}\hl{ }\hl{te}\hl{te}\hl{ }\hl{[SEP]} \textbf{\dataaug~prediction: True} 
 & \definecolor{highlight}{RGB}{163,189,219}\sethlcolor{highlight}\hl{[CLS]}\xspace\xspace\xspace...\xspace\xspace
 \hl{ }\hl{in}\hl{ }\hl{may}\hl{ }\hl{1904}\hl{ }\hl{,}\hl{ }\hl{the}\hl{ }\hl{couple}\hl{ }\hl{'}\hl{ }\hl{s}\hl{ }\hl{first}\hl{ }\hl{son}\hl{ }\hl{,}\hl{ }\hl{hans}\hl{ }\hl{albert}\hl{ }\hl{einstein}\hl{ }\hl{,}\hl{ }\hl{was}\hl{ }\hl{born}\hl{ }\hl{in}\hl{ }\hl{bern}\hl{ }\hl{,}\hl{ }\hl{switzerland}\hl{ }\hl{.}
 \hl{ }\hl{their}\hl{ }\hl{second}\hl{ }\hl{son}\hl{ }\hl{,}\hl{ }\hl{eduard}\hl{ }\hl{,}\hl{ }\hl{was}\hl{ }\hl{born}\hl{ }\hl{in}\hl{ }\hl{zurich}\hl{ }\hl{in}\hl{ }\hl{july}\hl{ }\hl{1910}\hl{ }\hl{.}\hl{ }\hl{in}\hl{ }\hl{1914}\hl{ }\hl{,}\hl{ }\hl{the}\hl{ }\hl{couple}\hl{ }\hl{separated}\hl{ }\hl{;}\hl{ }\hl{einstein}\hl{ }\hl{moved}\hl{ }\hl{to}\hl{ }\hl{berlin}\hl{ }\hl{and}\hl{ }\hl{his}\hl{ }\hl{wife}\hl{ }\hl{remained}\hl{ }\hl{in}\hl{ }\hl{zurich}\hl{ }\hl{with}\hl{ }\hl{their}\hl{ }\hl{sons}\hl{ }\hl{.}\hl{ }\hl{they}\hl{ }\hl{divorced}\hl{ }\hl{on}\hl{ }\hl{14}\hl{ }\hl{february}\hl{ }\hl{1919}\hl{ }\hl{,}\hl{ }\hl{having}\hl{ }\hl{lived}\hl{ }\hl{apart}\hl{ }\hl{for}\hl{ }\hl{five}\hl{ }\hl{years}\hl{ }\hl{.}\xspace\xspace\xspace...\xspace\xspace a - tete did n ' t stay in basel after charles and houben separated .\xspace\xspace\xspace...\xspace\xspace\hl{ }\hl{[SEP]}\hl{ }\hl{who}\hl{ }\hl{did}\hl{ }\hl{n}\hl{ }\hl{'}\hl{ }\hl{t}\hl{ }\hl{stay}\hl{ }\hl{in}\hl{ }\hl{zurich}\hl{ }\hl{after}\hl{ }\hl{albert}\hl{ }\hl{and}\hl{ }\hl{mari}\hl{c}\hl{ }\hl{separated}\hl{ }\hl{?}\hl{ }\hl{|}\hl{ }\hl{|}\hl{ }\hl{te}\hl{te}\hl{ }\hl{[SEP]} \textbf{\dataaugwithnatokens~prediction: False} 
 & \definecolor{highlight}{RGB}{163,189,219}\sethlcolor{highlight}\hl{[CLS]}\xspace\xspace\xspace...\xspace\xspace 
 in may 1904 , the couple ' s first son , hans albert einstein , was born in bern , switzerland . 
 their second son , eduard , was born in zurich in july 1910 \hl{.}\hl{ }\hl{in}\hl{ }\hl{1914}\hl{ }\hl{,}\hl{ }\hl{the}\hl{ }\hl{couple}\hl{ }\hl{separated}\hl{ }\hl{;}\hl{ }\hl{einstein}\hl{ }\hl{moved}\hl{ }\hl{to}\hl{ }\hl{berlin}\hl{ }\hl{and}\hl{ }\hl{his}\hl{ }\hl{wife}\hl{ }\hl{remained}\hl{ }\hl{in}\hl{ }\hl{zurich}\hl{ }\hl{with}\hl{ }\hl{their}\hl{ }\hl{sons}\hl{ }\hl{.}\hl{ }\hl{they}\hl{ }\hl{divorced}\hl{ }\hl{on}\hl{ }\hl{14}\hl{ }\hl{february}\hl{ }\hl{1919}\hl{ }\hl{,}\hl{ }\hl{having}\hl{ }\hl{lived}\hl{ }\hl{apart}\hl{ }\hl{for}\hl{ }\hl{five}\hl{ }\hl{years}\hl{ }\hl{.}\xspace\xspace\xspace...\xspace\xspace a - tete did n ' t stay in basel after charles and houben separated .\xspace\xspace\xspace...\xspace\xspace \hl{[SEP]}\hl{ }\hl{who}\hl{ }\hl{did}\hl{ }\hl{n}\hl{ }\hl{'}\hl{ }\hl{t}\hl{ }\hl{stay}\hl{ }\hl{in}\hl{ }\hl{zurich}\hl{ }\hl{after}\hl{ }\hl{albert}\hl{ }\hl{and}\hl{ }\hl{mari}\hl{c}\hl{ }\hl{separated}\hl{ }\hl{?}\hl{ }\hl{|}\hl{ }\hl{|}\hl{ }\hl{te}\hl{te}\hl{ }\hl{[SEP]}    \textbf{\dataaugwithhuman~prediction: False}
\\ \midrule
\multicolumn{3}{c}{\textbf{(B) Example 2}, true label: True}          \\ \midrule
\definecolor{highlight}{RGB}{244,236,164}\sethlcolor{highlight}\hl{[CLS]}\xspace\xspace\xspace...\xspace\xspace \hl{on}\hl{ }\hl{the}\hl{ }\hl{day}\hl{ }\hl{of}\hl{ }\hl{the}\hl{ }\hl{party}\hl{ }\hl{,}\hl{ }\hl{all}\hl{ }\hl{five}\hl{ }\hl{friends}\hl{ }\hl{showed}\hl{ }\hl{up}\hl{ }\hl{.}\hl{ }\hl{each}\hl{ }\hl{friend}\hl{ }\hl{had}\hl{ }\hl{a}\hl{ }\hl{present}\hl{ }\hl{for}\hl{ }\hl{susan}\hl{ }\hl{.} \definecolor{highlight}{RGB}{238,186,180}\sethlcolor{highlight} \hl{6}\hl{ }\hl{thank}\hl{ }\hl{-}\hl{ }\hl{you}\hl{ }\hl{cards}\hl{ }\hl{did}\hl{ }\hl{helen}\hl{ }\hl{send}\hl{ }\hl{.} \definecolor{highlight}{RGB}{244,236,164} \hl{susan}\hl{ }\hl{was}\hl{ }\hl{happy}\hl{ }\hl{and}\hl{ }\hl{sent}\hl{ }\hl{each}\hl{ }\hl{friend}\hl{ }\hl{a}\hl{ }\hl{thank}\hl{ }\hl{you}\hl{ }\hl{card}\hl{ }\hl{the}\hl{ }\hl{next}\hl{ }\hl{week}\hl{ }\hl{.}\hl{ }\hl{[SEP]}\hl{ }\hl{how}\hl{ }\hl{many}\hl{ }\hl{thank}\hl{ }\hl{-}\hl{ }\hl{you}\hl{ }\hl{cards}\hl{ }\hl{did}\hl{ }\hl{susan}\hl{ }\hl{send}\hl{ }\hl{?}\hl{ }\hl{|}\hl{ }\hl{|}\hl{ }\hl{5}\hl{ }\hl{[SEP]} \textbf{\dataaug~prediction: False} 
& \definecolor{highlight}{RGB}{163,189,219}\sethlcolor{highlight}\hl{[CLS]}\xspace\xspace\xspace...\xspace\xspace\hl{ }\hl{on}\hl{ }\hl{the}\hl{ }\hl{day}\hl{ }\hl{of}\hl{ }\hl{the}\hl{ }\hl{party}\hl{ }\hl{,}\hl{ }\hl{all}\hl{ }\hl{five}\hl{ }\hl{friends}\hl{ }\hl{showed}\hl{ }\hl{up}\hl{ }\hl{.}\hl{ }\hl{each}\hl{ }\hl{friend}\hl{ }\hl{had}\hl{ }\hl{a}\hl{ }\hl{present}\hl{ }\hl{for}\hl{ }\hl{susan}\hl{ }\hl{.} 6 thank - you cards did helen send . \hl{susan}\hl{ }\hl{was}\hl{ }\hl{happy}\hl{ }\hl{and}\hl{ }\hl{sent}\hl{ }\hl{each}\hl{ }\hl{friend}\hl{ }\hl{a}\hl{ }\hl{thank}\hl{ }\hl{you}\hl{ }\hl{card}\hl{ }\hl{the}\hl{ }\hl{next}\hl{ }\hl{week}\hl{ }\hl{.}\hl{ }\hl{[SEP]}\hl{ }\hl{how}\hl{ }\hl{many}\hl{ }\hl{thank}\hl{ }\hl{-}\hl{ }\hl{you}\hl{ }\hl{cards}\hl{ }\hl{did}\hl{ }\hl{susan}\hl{ }\hl{send}\hl{ }\hl{?}\hl{ }\hl{|}\hl{ }\hl{|}\hl{ }\hl{5}\hl{ }\hl{[SEP]} \textbf{\dataaugwithnatokens~prediction: True} 
& \definecolor{highlight}{RGB}{163,189,219}\sethlcolor{highlight}\hl{[CLS]}\xspace\xspace\xspace...\xspace\xspace on the day of the party , all five friends showed up . each friend had a present for susan . 6 thank - you cards did helen send . \hl{susan}\hl{ }\hl{was}\hl{ }\hl{happy}\hl{ }\hl{and}\hl{ }\hl{sent}\hl{ }\hl{each}\hl{ }\hl{friend}\hl{ }\hl{a}\hl{ }\hl{thank}\hl{ }\hl{you}\hl{ }\hl{card}\hl{ }\hl{the}\hl{ }\hl{next}\hl{ }\hl{week}\hl{ }\hl{.}\hl{ }\hl{[SEP]}\hl{ }\hl{how}\hl{ }\hl{many}\hl{ }\hl{thank}\hl{ }\hl{-}\hl{ }\hl{you}\hl{ }\hl{cards}\hl{ }\hl{did}\hl{ }\hl{susan}\hl{ }\hl{send}\hl{ }\hl{?}\hl{ }\hl{|}\hl{ }\hl{|}\hl{ }\hl{5}\hl{ }\hl{[SEP]} \textbf{\dataaugwithhuman~prediction: False} \\
\midrule
\multicolumn{3}{c}{\textbf{(C) Example 3}, true label: False}      \\ \midrule
\definecolor{highlight}{RGB}{244,236,164 }\sethlcolor{highlight}\hl{[CLS]}\xspace\xspace\xspace...\xspace\xspace \hl{roman}\hl{ }\hl{legions}\hl{ }\hl{encountered}\hl{ }\hl{the}\hl{ }\hl{stronghold}\hl{s}\hl{ }\hl{of}\hl{ }\hl{the}\hl{ }\hl{castle}\hl{ }\hl{rock}\hl{ }\hl{and}\hl{ }\hl{arthur}\hl{ }\hl{'}\hl{ }\hl{s}\hl{ }\hl{seat}\hl{ }\hl{,}\hl{ }\hl{held}\hl{ }\hl{by}\hl{ }\hl{a}\hl{ }\hl{tribe}\hl{ }\hl{of}\hl{ }\hl{ancient}\hl{ }\hl{brit}\hl{ons}\hl{ }\hl{known}\hl{ }\hl{as}\hl{ }\hl{the}\hl{ }\hl{vo}\hl{tad}\hl{ini}\hl{ }\hl{.} \definecolor{highlight}{RGB}{238,186,180}
\hl{the}\hl{ }\hl{mer}\hl{cian}\hl{s}\hl{ }\hl{were}\hl{ }\hl{probably}\hl{ }\hl{the}\hl{ }\hl{ancestors}\hl{ }\hl{of}\hl{ }\hl{the}\hl{ }\hl{mana}\hl{w}\hl{ }\hl{.} \definecolor{highlight}{RGB}{244,236,164 }\sethlcolor{highlight} \hl{little}\hl{ }\hl{is}\hl{ }\hl{recorded}\hl{ }\hl{about}\hl{ }\hl{this}\hl{ }\hl{group}\hl{ }\hl{,}\hl{ }\hl{but}\hl{ }\hl{they}\hl{ }\hl{were}\hl{ }\hl{probably}\hl{ }\hl{the}\hl{ }\hl{ancestors}\hl{ }\hl{of}\hl{ }\hl{the}\hl{ }\hl{god}\hl{od}\hl{din}\hl{ }\hl{,}\hl{ }\hl{whose}\hl{ }\hl{feat}\hl{s}\hl{ }\hl{are}\hl{ }\hl{told}\hl{ }\hl{in}\hl{ }\hl{a}\hl{ }\hl{seventh}\hl{ }\hl{-}\hl{ }\hl{century}\hl{ }\hl{old}\hl{ }\hl{welsh}\hl{ }\hl{manuscript}\hl{ }\hl{.}\xspace\xspace\xspace...\xspace\xspace the god\xspace\xspace\xspace...\xspace\xspace din\xspace\xspace\xspace...\xspace\xspace \hl{[SEP]}\hl{ }\hl{who}\hl{ }\hl{were}\hl{ }\hl{probably}\hl{ }\hl{the}\hl{ }\hl{ancestors}\hl{ }\hl{of}\hl{ }\hl{the}\hl{ }\hl{god}\hl{od}\hl{din}\hl{ }\hl{?}\hl{ }\hl{|}\hl{ }\hl{|}\hl{ }\hl{the}\hl{ }\hl{pic}\hl{ts}\hl{ }\hl{[SEP]}  \textbf{\dataaug~prediction: True}              
& \definecolor{highlight}{RGB}{163,189,219}\sethlcolor{highlight}\hl{[CLS]}\xspace\xspace\xspace...\xspace\xspace\hl{ }\hl{roman}\hl{ }\hl{legions}\hl{ }\hl{encountered}\hl{ }\hl{the}\hl{ }\hl{stronghold}\hl{s}\hl{ }\hl{of}\hl{ }\hl{the}\hl{ }\hl{castle}\hl{ }\hl{rock}\hl{ }\hl{and}\hl{ }\hl{arthur}\hl{ }\hl{'}\hl{ }\hl{s}\hl{ }\hl{seat}\hl{ }\hl{,}\hl{ }\hl{held}\hl{ }\hl{by}\hl{ }\hl{a}\hl{ }\hl{tribe}\hl{ }\hl{of}\hl{ }\hl{ancient}\hl{ }\hl{brit}\hl{ons}\hl{ }\hl{known}\hl{ }\hl{as}\hl{ }\hl{the}\hl{ }\hl{vo}\hl{tad}\hl{ini}\hl{ }\hl{.} the mercians were probably the ancestors of the manaw . \hl{little}\hl{ }\hl{is}\hl{ }\hl{recorded}\hl{ }\hl{about}\hl{ }\hl{this}\hl{ }\hl{group}\hl{ }\hl{,}\hl{ }\hl{but}\hl{ }\hl{they}\hl{ }\hl{were}\hl{ }\hl{probably}\hl{ }\hl{the}\hl{ }\hl{ancestors}\hl{ }\hl{of}\hl{ }\hl{the}\hl{ }\hl{god}\hl{od}\hl{din}\hl{ }\hl{,}\hl{ }\hl{whose}\hl{ }\hl{feat}\hl{s}\hl{ }\hl{are}\hl{ }\hl{told}\hl{ }\hl{in}\hl{ }\hl{a}\hl{ }\hl{seventh}\hl{ }\hl{-}\hl{ }\hl{century}\hl{ }\hl{old}\hl{ }\hl{welsh}\hl{ }\hl{manuscript}\hl{ }\hl{.}\xspace\xspace\xspace...\xspace\xspace\hl{ }\hl{the}\hl{ }\hl{god}\xspace\xspace\xspace...\xspace\xspace\hl{din}\xspace\xspace\xspace...\xspace\xspace\hl{ }\hl{[SEP]}\hl{ }\hl{who}\hl{ }\hl{were}\hl{ }\hl{probably}\hl{ }\hl{the}\hl{ }\hl{ancestors}\hl{ }\hl{of}\hl{ }\hl{the}\hl{ }\hl{god}\hl{od}\hl{din}\hl{ }\hl{?}\hl{ }\hl{|}\hl{ }\hl{|}\hl{ }\hl{the}\hl{ }\hl{pic}\hl{ts}\hl{ }\hl{[SEP]}  \textbf{\dataaugwithnatokens~prediction: True}
& \definecolor{highlight}{RGB}{163,189,219}\sethlcolor{highlight}\hl{[CLS]}\xspace\xspace\xspace...\xspace\xspace roman legions encountered the strongholds of the castle rock and arthur ' s seat , held by a tribe of ancient britons known as the votadini . \hl{the}\hl{ }\hl{mer}\hl{cian}\hl{s}\hl{ }\hl{were}\hl{ }\hl{probably}\hl{ }\hl{the}\hl{ }\hl{ancestors}\hl{ }\hl{of}\hl{ }\hl{the}\hl{ }\hl{mana}\hl{w}\hl{ }\hl{.}\hl{ }\hl{little}\hl{ }\hl{is}\hl{ }\hl{recorded}\hl{ }\hl{about}\hl{ }\hl{this}\hl{ }\hl{group}\hl{ }\hl{,}\hl{ }\hl{but}\hl{ }\hl{they}\hl{ }\hl{were}\hl{ }\hl{probably}\hl{ }\hl{the}\hl{ }\hl{ancestors}\hl{ }\hl{of}\hl{ }\hl{the}\hl{ }\hl{god}\hl{od}\hl{din}\hl{ }\hl{,}\hl{ }\hl{whose}\hl{ }\hl{feat}\hl{s}\hl{ }\hl{are}\hl{ }\hl{told}\hl{ }\hl{in}\hl{ }\hl{a}\hl{ }\hl{seventh}\hl{ }\hl{-}\hl{ }\hl{century}\hl{ }\hl{old}\hl{ }\hl{welsh}\hl{ }\hl{manuscript}\hl{ }\hl{.}\xspace\xspace\xspace...\xspace\xspace 
\hl{the}\hl{ }\hl{god}\xspace\xspace\xspace...\xspace\xspace\hl{din}\xspace\xspace\xspace...\xspace\xspace\hl{ }\hl{[SEP]}\hl{ }\hl{who}\hl{ }\hl{were}\hl{ }\hl{probably}\hl{ }\hl{the}\hl{ }\hl{ancestors}\hl{ }\hl{of}\hl{ }\hl{the}\hl{ }\hl{god}\hl{od}\hl{din}\hl{ }\hl{?}\hl{ }\hl{|}\hl{ }\hl{|}\hl{ }\hl{the}\hl{ }\hl{pic}\hl{ts}\hl{ }\hl{[SEP]} \textbf{\dataaugwithhuman~prediction: False} 
\\ \bottomrule
\end{tabular}
\caption{Example outputs from \dataaugwithnatokens and \dataaugwithhuman with BERT in \multirc. 
Attack tokens are marked in \hlattack{red}. True human rationales are marked in \hlhuman{yellow}, 
and predicted rationales are marked in \hlsupervision{blue}. We only show tokens where generated rationales disagree with each other or with the \hr/attack.}
\label{tab:example_table}
\end{table*}

%% file: discussion.tex
\section{Concluding Discussion}

In this study, we find that adding 
an explicit extractor layer helps a 
model learn to ignore additive adversarial attacks produced by the \addsent attack more effectively than conventional adversarial training via data augmentation. 

This is an exciting result because it defeats an attack which is otherwise stubbornly effective against even copious adversarial data augmentation. It is a novel use for this type of explicit token relevance representation, which is more typically applied for model interpretability \citep{lei-etal-2016-rationalizing}. This makes it related to defenses like \citet{cohen_certified_2019} which allow the model to reject inputs as out-of-distribution and abstain from prediction, but it differs in rejecting only part of the input, making a prediction from the remainder as usual.

\para{Generality.} As \citet{carlini_evaluating_2019} note, it is easy to overstate claims in evaluating adversarial defenses. Hence, we note that our results pertain only to the \addsent attack, and perform favorably 
against a baseline defense in adversarial training via data augmentation. 
Since most adversarial training approaches assume the ability to {\em generate} a large number of synthetic attack examples, it is reasonable to further assume that we have access to the positions of the attacks.
However, such knowledge about attacks may not be available in general.
Nevertheless, the success of the rationale model architecture in learning to occlude adversarial tokens does hold promise for a more general defense based on a wider range of possible attacks and possible defenses by the extractor layer.

\para{Utility of human rationales.} We explore the possibility that human-provided explanations may make the model more robust against adversarial attacks. We mostly find that they do not, with the notable exception of \bert on \multirc, the only case in which the augmentation brings the model back to clean test accuracy.
While it does provide an advantage of sparsity over supervision with non-attack tokens, this advantage alone may not justify the cost of collecting human explanations for robustness.
Further understanding of human rationales and novel learning strategies are required for improving model robustness.

\para{Future directions.} A generalization of our approach might convert the  ``extractor'' layer into a more general ``defender'' layer capable of issuing a wider range of corrections in response to a wider range of attacks. It could, for example, learn to defend against attacks based on input removal (e.g. \citet{feng_pathologies_2018}) by training to recognize gaps in the input and fill them via generative closure. This defender could be coupled with a self-supervision style approach (e.g., \citet{hendrycks_using_2019}) involving an ``attacker'' capable of levying various types of attack against the model. 
We leave such a generalization for future work.


%% file: appendix.tex

\input{tables/hyperparameters}

\section{Design choices and implementation details}
\label{appendix:implementation}

We use the HuggingFace \citep{wolf_huggingfaces_2020} distributions of BERT and RoBERTa, and Pytorch Lightning \citep{falcon_pytorch_2019} for model training. Models are trained for a minimum of 3 epochs with early stopping based on a patience of 5 validation intervals, evaluated every 0.2 epochs.

In practice, we find it useful to pretrain the predictor layer $f$ of the rationale model on full input before jointly training it with the extractor $g$. We observe that this trick stabilizes training and helps prevent mode collapse. In producing the predicted rationale, we automatically assign a 1 (indicating relevance) to every token in the query, so that they are always fully visible to the predictor and the effect of the extractor is in adjudicating which tokens of the document are used or ignored. 



Traditionally, this style of rationale model produces binary predicted rationales via either reinforcement learning \cite{williams_simple_1992} or categorical reparameterization such as Gumbel Softmax \citep{jang_categorical_2016}. One argument for this approach is that binary rationales are more interpretable, leaving less ambiguity about the precise role of a given token in the model's output. Another argument is that transformer-based models like BERT don't have a native interpretation for partially-masked input, whereas fully-masked input can represent in-distribution modifications such as the [MASK] token substitution used in masked-LM pretraining. 

However, we find that relaxing this binary constraint leads to better outcomes for adversarial training. Thus, our model produces predicted rationale $\hat{\vect{r}}$ by passing predicted rationale logits $\vect{\phi}$ through a sigmoid function. The masking function $m$ we use is simply to multiplicatively weight $\vect{x}$ by predicted rationale $\hat{\vect{r}}$ during training (we discretize $\vect{r}$ during testing),

\begin{equation}
m(\vect{x}, \hat{\vect{r}}) = \hat{\vect{r}} \cdot \vect{x}
\end{equation}

From a theoretical perspective, jointly optimizing the rationale extractor $g$ and label predictor $f$ should allow the model to predict rationale $\hat{\vect{r}}$ that is more adapted to the predictor. Separately optimizing both components implies that the rationale extractor does not get penalized for poor label prediction performance, and often leads to predicted rationale that is closer to human rationale $\vect{r}$. In our experiments, we include both training setups as a hyperparameter.



\section{Hyperparameters}
For our experiments, we fine-tune both the rationale extractor $g$ and predictor $f$ for the rationale models from a pretrained language model. We finetune \bert components from a pre-trained \textit{bert-base-uncased} model, and \roberta from a pre-trained \textit{roberta-large} model. We use an Adam optimizer with with $\beta_1 = 0.9$ and 
$\beta_2 = 0.999$ for all experiments.

We find gradient accumulation helps with training stability of \bert and \roberta, and we report gradient accumulation as a hyperparameter for both models. Table \ref{tab:hyper} describes a list of hyperparameters we use for both \bert and \roberta. We do a grid search over all combinations of hyperparameters listed in table \ref{tab:hyper}, and we report results of the model that achieves the highest performance on the original dev set.

\section{Computation Details}

We ran our experiments on a mix of RTX 3090, A30 and A40 GPUs. All experiments combined take less than 300 GPU hours.

The rationale model has about two times the parameters of its base model. The \bert-based rationale model has 220 million parameters and \roberta-based rationale model has 708 million parameters. Both models can be trained on a GPU with 24GB of memory.
Training the rationale model typically takes double the training time compared to the standard model.
On a RTX 3090 GPU, training a BERT Rationale model for SQuAD takes about 4 hours, while training the standard BERT model takes 2.5 hours.

\section{Statistical Significance}

\input{tables/significance.tex}

Due to limited computational resources and a large number of experiment conditions,
our experiments are not repeated across multiple random seeds.
To verify the statistical significance of improvements of the top-performing rationale
models against the strongest baselines, we report Wilcoxon signed-rank test results
in \tableref{tab:stat-sig}. Note that we do not report \squad results due to the
incompatibility of the statistical test with the metric (Span-F1).
We find that for 3 out of 4 model-dataset pairs, the observed improvements of
the rationale models over the baselines are highly significant.

\section{More Error Analysis}
\textbf{Easy examples have high jaccard similarity between \hr and \qa.} All three models excel at these examples.
High similarity should help models to find \hr or generate rationales that mimic \hr easily, but we also observe that the generated rationales do not necessarily provide the greatest alignment with \hr for examples \br models get correct. 
For instance, rationale F1 is 53.9 for examples that \brlr gets correct and \bc gets wrong, which is smaller than rationale F1 (56.2) for examples both models get wrong.
Notice that attack and \hr are similar due to the attack generation technique, but this does not affect model performance because training with augmentation allows the rationale models to ignore attack tokens (attack recall = 89.3 and 97.4 for \br models).
Likewise, we think \bcaug also benefits from the high similarity to identify important text areas and learns to ignore attacks from training augmentation.

\textbf{\br models handle denser \hr slightly better than \bcaug.} 
We define sparsity of $\mathbf{X}$ as the number of tokens in  $\mathbf{X}$ divided by the total number of tokens in the input, so larger sparsity correspond to dense rationales.
Counter-intuitively, all three models are bad at examples with the most dense \hr. This can be accounted for by the fact that these are also examples where \qa and \hr have the least jaccard similarity: \hr sparsity and the jaccard similarity has a Pearson's coefficient of 0.25 (p < 0.001).
Thus, examples with denser \hr are likely to contain confusing information for models. 
We find \br models can resist this confusion better than \bcaug. For instance, \hr sparsity = 0.167 when \brlr is correct bu \bc is wrong, and it is 0.165 when \bc is correct but \br is wrong. 
\input{venn.tex}
\input{tables/brlr-vs-brlna}


\section{More Results on \squad}
\input{squad}

%% file: tables/hyperparameters.tex
\begin{table*}[]
\centering
\begin{tabular}{@{}l|l|l@{}}
Parameter                          & BERT Rationale               & RoBERTa Rationale            \\ \midrule
Batch Size                         & 8                            & 8                            \\
Learning Rate                      & 2e-5                         & 5e-6                         \\
Gradient Accumulation              & 10 batches                   & 8 batches                    \\
Masking Strategy $m$                   & $m_{\text{zero}}$, $m_{\text{mask}}$ & $m_{\text{zero}}$, $m_{\text{mask}}$ \\
Prediction Supervision Loss Weight & 1.0                          & 1.0                          \\
Rationale Supervision Loss Weight $\lambda_1$  & 1.0                          & 1.0                          \\
Sparsity Loss Weight $\lambda_2$               & 0.0, 0.1, 0.2, 0.3           & 0.0, 0.1, 0.2, 0.3           \\
Jointly Optimized       & True, False                  & True, False                 
\end{tabular}
\caption{Hyperparameters used in parameter search and training. 
}
\label{tab:hyper}
\end{table*}

%% file: tables/significance.tex
\begin{table*}[t]
\centering
\begin{tabular}{@{}l|l|l|l|l@{}}
Model & Dataset & Baseline Variant (Acc) & Rationale Variant (Acc) & p-value \\ \midrule
\bert & \multirc & \dataaugtenx ($65.9\%$) & \dataaugwithhuman ($69.4\%$) & $1.0 \times 10^{-4}$ \\
\bert & \fever & \dataaug ($84.8\%$) & \dataaugwithnatokens ($87.7\%$) & $8.1\ \times 10^{-13}$ \\
\roberta & \multirc & \dataaug ($84.8\%$) & \dataaugwithnatokens ($85.1\%$) & $9.4 \times 10^{-4}$ \\
\roberta & \fever & \dataaugtenx ($93.4\%$) & \dataaugwithnatokens ($93.2\%$) & $0.18$ \\
\end{tabular}
\caption{Wilcoxon signed-rank test for statistical significance of improvements of the
top-performing rationale models over the strongest baselines across models and
datasets.}
\label{tab:stat-sig}
\end{table*}

%% file: venn.tex
\begin{figure}[]
  \centering
  \includegraphics[keepaspectratio, width=0.4\textwidth]{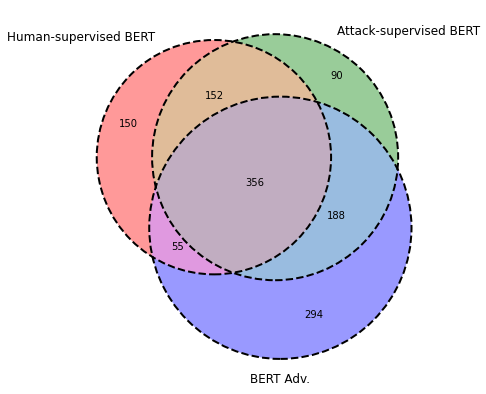}
  \caption{Venn diagram for errors by \bcaug, \brlr, and \brlna. 
  }
  \label{fig:venn} 
\end{figure}

%% file: tables/brlr-vs-brlna.tex
\begin{table*}[t]
\centering
\begin{tabular}{lllll}
\toprule
             & \begin{tabular}[c]{@{}l@{}}Input Length\end{tabular} & \begin{tabular}[c]{@{}l@{}}Human Rationale Length\end{tabular} &  &  \\ \midrule
\brlr correct, \\ \brlna wrong & 357.097                                                                            & 360.278                                                                            &  &  \\ \midrule
\brlna correct, \\ \brlr wrong  & 81.191                                                                             & 79.098                                                                             &  &  \\ \bottomrule
             &                                                                                    &                                                                                    &  & 
\end{tabular}
\caption{Input and \hr length of mistakes by \brlna and \brlr.} 
\label{tab:brlr-vs-brlna} 
\end{table*}


%% file: squad.tex

\input{tables/squad_results}


In Table \ref{tab:squad-results}, we report the F1 scores of \bert Classification and \bert Rationale models on four different evaluation sets: the original \squad development set dev, the synthetic attack set \addsentsynthetic, the human re-written and filtered attack set \addsenthuman and the human-generated, model-free attack baseline \addonesenthuman.

Similar to \secref{sec:main_results}, we find the performances on the clean set (\squad dev) to be approximately equal across models and training schemes. We observe a slight drop (-0.4\%) on dev accuracy when adding adversarial training to the \bert classification, which points to compromised learning on the original \squad task after adding adversarial examples. Without adversarial training, we observe roughly 38\%, 24\%, 15\% performance decreases for \addsentsynthetic, \addsenthuman, \addonesenthuman, respectively. All three attacks lead to much more significant performance drops than the \addsent attack on \multirc and \fever, which yields an approximate 6\% performance drop. This observation is likely due to the differences in how the tasks are formulated across datasets: it is plausible that an additive attack such as \addsent is more effective on a span-extraction style QA task (\squad) than on answer classification style QA tasks (\multirc and \fever).

Surprisingly, the synthetic attack \addsentsynthetic is more effective than human generated \addsenthuman prior to adversarial training. Since the \addsent attack works by mutating the query and adding a fake answer, the synthetic attack often appears syntactically similar to the query. On the other hand, human generated attacks in \addsenthuman often fits more naturally in the document and grammatically correct, but does not mirror the structure of the query. For a model that solves the QA task by simply looking for the best match of the query inside a document while skipping complex reasoning, it's conceivable that \addsentsynthetic leads to the greatest performance drop.

Since \addsenthuman and \addonesenthuman are attack examples re-written and filtered by humans, we use them as a proxy for understanding the model behavior in a real-world setting. We find the \bert Rationale model with attack rationale supervision significantly outperforms the \bert Classification baseline trained with adversarial augmentation (+2.7\% on \addsenthuman, +2.2\% on \addonesenthuman). Similar to findings in \secref{sec:main_results}, we observe attack rationale supervision (\dataaugwithnatokens) as a more effective adversarial training method than adversarial data augmentation (\dataaug). It is worth noting that the despite training on the synthetic attacks, the rationale model demonstrates strong ability to generalize knowledge learned from synthetic attacks to tune out human-rewritten attacks, which explains the strong performance on \addsenthuman and \addonesenthuman.

An apparent anomaly in Table \ref{tab:squad-results} is the strong performance of \bert Classification on \addsentsynthetic (93.3\%), which is even greater than the performance on the clean development set (86.0\%). During the \addsent attack, the answer is mutated into an incorrect, but similar phrase (e.g. Dallas Cowboys $\to$ Michigan Vikings). The presence of a mutated answer in the passage likely gives the model additional information on what the correct answer looks like, while the rationale model avoids utilizing this information to a much higher degree (88.3\%) due to attack rationale supervision. This "mutated answer" signal is akin to spurious correlations in datasets, and our method helps the \bert rationale model ignore such spurious correlations a lot more effectively than the baseline \bert model.

Overall, these analyses shine light on the benefits of including rationale supervision in adversarial training. Our method achieves greater adversarial robustness in a close-to-real-world setting (\addsenthuman and \addonesenthuman) by generalizing from synthetic attacks.

%% file: tables/squad_results.tex
\begin{table*}[t]
\centering
\begin{tabular}{llcccc}
\hline
Architecture & Training & dev & \addsentsynthetic & \addsenthuman & \addonesenthuman \\ \hline
\multirow{2}{*}{Bert Classification} & \nodataaug           & 86.4 & 48.7 & 62.8 & 71.2 \\
                                     & \dataaug             & 86.0 & 93.3 & 80.4 & 81.9 \\
                                     & \dataaugtenx             & 82.2 & \textbf{95.9} & 78.0 & 79.7 \\
                                     \hline
\multirow{2}{*}{Bert Rationale}      & \nodataaug           & \textbf{86.6} & 47.7 & 62.0 & 70.4 \\
                                     & \dataaugwithnatokens & 86.5 & 88.3 & \textbf{83.1} & \textbf{84.1} \\ \hline
\end{tabular}
\caption{F1 scores of \bert Rationale and Classification models on the \squad task.}
\label{tab:squad-results}
\end{table*}